\documentclass[11pt,a4paper]{article}
\usepackage[hyperref]{acl2019}
\usepackage{times}
\usepackage{latexsym}
\usepackage{url}
\usepackage{verbatim}
\usepackage{fancyvrb}
\usepackage{adjustbox}

\usepackage{xr}

\makeatletter
\newcommand*{\addFileDependency}[1]{
  \typeout{(#1)}
  \@addtofilelist{#1}
  \IfFileExists{#1}{}{\typeout{No file #1.}}
}
\makeatother
 
\newcommand*{\myexternaldocument}[1]{%
    \externaldocument{#1}%
    \addFileDependency{#1.tex}%
    \addFileDependency{#1.aux}%
}

\myexternaldocument{supplemental}

\usepackage{todonotes}

\aclfinalcopy 

\title{\textsc{DisSim}: A Discourse-Aware Syntactic Text Simplification Framework\\ for English and German}

\author{Christina Niklaus\textsuperscript{1}\textsuperscript{3}, Matthias Cetto\textsuperscript{1}, Andr\'{e} Freitas\textsuperscript{2}, \and Siegfried Handschuh\textsuperscript{1}\textsuperscript{3} \\
  \textsuperscript{1} University of St.Gallen\\
  {\small{{\tt \{christina.niklaus, matthias.cetto, siegfried.handschuh\}{\tt @unisg.ch}}}}\\
  \textsuperscript{2} University of Manchester\\
  {\small{{\tt andre.freitas@manchester.ac.uk}}}\\
  \textsuperscript{3} University of Passau\\
  {\small{{\tt \{christina.niklaus, siegfried.handschuh\}{\tt @uni-passau.de}}}}
\\}

\date{}

\begin{document}
\maketitle
\begin{abstract}
  We introduce \textsc{DisSim}, a discourse-aware sentence splitting framework for English and German whose goal is to transform syntactically complex sentences into an intermediate representation that presents a simple and more regular structure which is easier to process for downstream semantic applications. For this purpose, we turn input sentences into a two-layered semantic hierarchy in the form of core facts and accompanying contexts, while identifying the rhetorical relations that hold between them. 
  In that way, we preserve the coherence structure of the input and, hence, its interpretability for downstream tasks.
\end{abstract}

\section{Introduction}

We developed a syntactic text simplification (TS) approach that can be used as a preprocessing step to facilitate and improve the performance of a wide range of artificial intelligence (AI) tasks, such as Machine Translation, Information Extraction (IE) or Text Summarization. Since shorter sentences are generally better processed by natural language processing (NLP) systems \cite{Narayan2017}, the goal of our approach is to \textbf{break down a complex source sentence into a set of minimal propositions}, i.e. a sequence of sound, self-contained utterances, with each of them presenting a minimal semantic unit that cannot be further decomposed into meaningful propositions \cite{bast2013open}.


However, any sound and coherent text is not simply a loose arrangement of self-contained units, but rather a logical structure of utterances that are semantically connected \cite{siddharthan2014survey}.
Consequently, when carrying out syntactic simplification operations without considering discourse implications, 
the rewriting may easily result in a disconnected sequence of simplified sentences that lack important contextual information, making the text harder to interpret. Thus, in order to \textbf{preserve the coherence structure} and, hence, the interpretability of the input, we developed a discourse-aware TS approach based on Rhetorical Structure Theory (RST) \cite{mann1988rhetorical}. It establishes a contextual hierarchy between the split components, 
and identifies and classifies the semantic relationship that holds between them. In that way, a complex source sentence is turned into a so-called discourse tree, consisting of a \textbf{set of hierarchically ordered and semantically interconnected sentences that present a simplified syntax} which is easier to process for downstream semantic applications and may support a faster generalization in machine learning tasks. 


\section{System Description}

\begin{figure*}[!htbp]
	\centering
  \includegraphics[width=\textwidth]{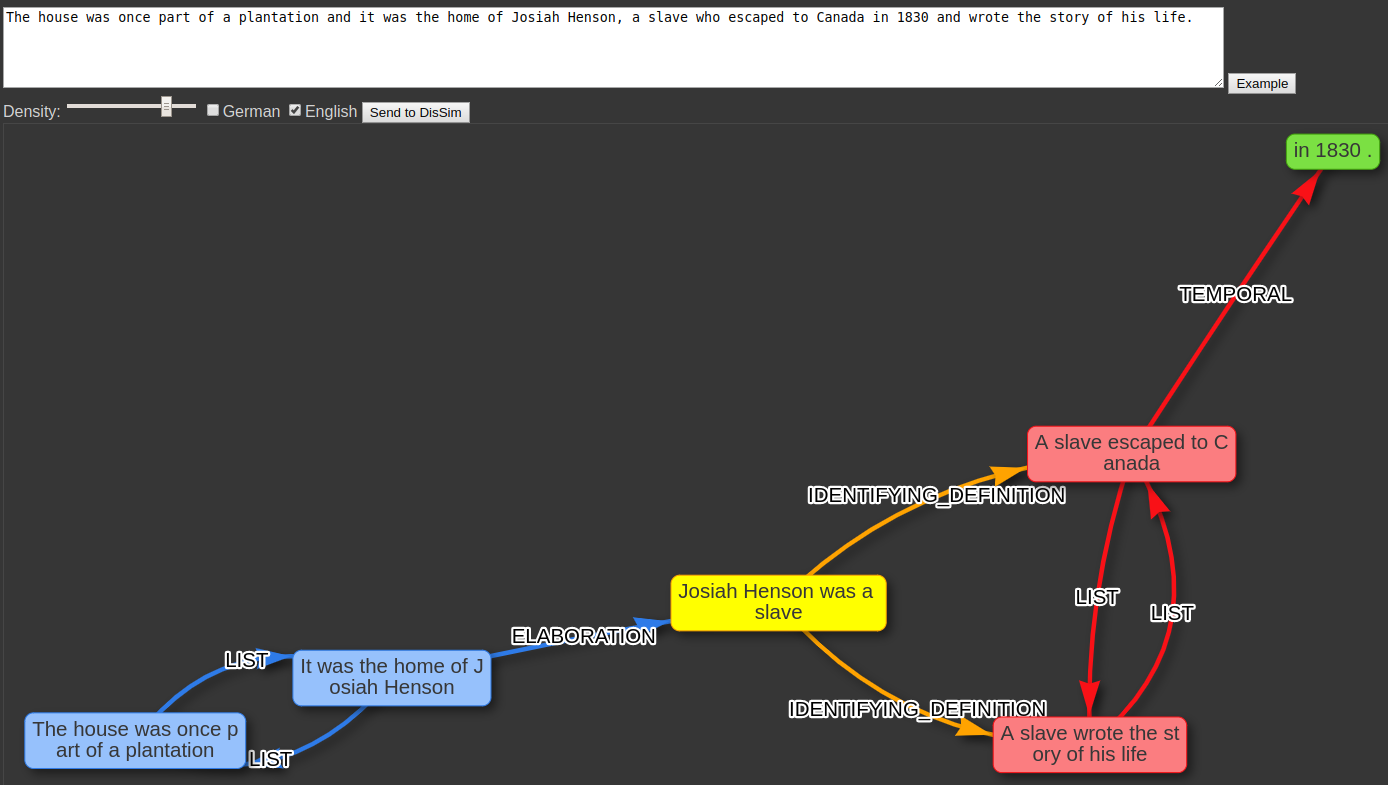}
	\caption{\textsc{DisSim}'s browser-based user interface. The simplified output is displayed in the form of a directed graph where the split sentences are connected by arrows whose labels denote the semantic relationship that holds between a pair of simplified sentences and whose direction indicates their contextual hierarchy. The colors signal different context layers. In that way, a semantic hierarchy of minimal, self-contained propositions is established.} 
	\label{screenshot_UI}
\end{figure*}

We present \textsc{DisSim}, a discourse-aware sentence splitting approach for English and German that creates a semantic hierarchy of simplified sentences.\footnote{The source code of our framework is available under \url{https://github.com/Lambda-3/DiscourseSimplification}.}
It takes a sentence as input and performs a recursive transformation process that is based upon a small set of 35 hand-crafted grammar rules for the English version and 29 rules for the German approach.\footnote{For reproducibility purposes, the complete set of transformation patterns is available under \url{https://github.com/Lambda-3/DiscourseSimplification/tree/master/supplemental_material}.} These patterns were heuristically determined in a comprehensive linguistic analysis 
and encode syntactic and lexical features that can be derived from a sentence's parse tree.\footnote{For the English version, we use Stanford's pre-trained lexicalized parser \cite{Socher2013}
to create a sentence's phrasal parse tree. For the German approach, we apply dependency parse structures generated by the spaCy parser {\scriptsize{(\url{https://spacy.io/})}}.} 
Each rule specifies (1) how to \textit{split up and rephrase} the input into structurally simplified sentences and (2) how to \textit{set up a semantic hierarchy} between them. 
They are recursively applied on a given source sentence in a top-down fashion. 
When no more rule matches, the algorithm stops and returns the generated discourse tree. 



\subsection{Split into Minimal Propositions}
In a first step, source sentences that present a complex linguistic form are turned into clean, compact structures by decomposing clausal and phrasal components. 
For this purpose, the transformation rules encode both the splitting points and rephrasing procedure for reconstructing proper sentences. 

\subsection{Establish a Semantic Hierarchy}
Each split will create two or more sentences with a simplified syntax. To establish a semantic hierarchy between them, two subtasks are carried out: 

\paragraph{Constituency Type Classification.}
First, we set up a contextual hierarchy between the split sentences by connecting them with information about their hierarchical level, similar to the concept of nuclearity in RST. For this purpose, we distinguish core sentences (\textit{nuclei}), which carry the key information of the input, from accompanying contextual sentences (\textit{satellites}) that disclose additional information about it. To differentiate between those two types of constituents, the transformation patterns encode a simple syntax-based approach where subordinate clauses/phrases are classified as context sentences, while superordinate as well as coordinate clauses/phrases are labelled as core. 



\paragraph{Rhetorical Relation Identification.}
Second, we aim to restore the semantic relationship between the disembedded components. For this purpose, we identify and classify the rhetorical relations that hold between the simplified sentences, making use of both syntactic features, which are derived from the input's parse tree structure, and lexical features in the form of cue phrases. Following the work of \newcite{Taboada13}, they are mapped to a predefined list of rhetorical cue words to infer the type of rhetorical relation.

\begin{figure*}[!ht]
\centering
\scriptsize
\begin{adjustbox}{max width=0.73\linewidth}
\begin{BVerbatim}[fontsize=\scriptsize]
Input sentence:
A fluoroscopic study known as an upper gastrointestinal series is typically the next step in management,
although if volvulus is suspected, caution with non water soluble contrast is mandatory as the usage of
barium can impede surgical revision and lead to increased post operative complications.

Supervised-OIE (alone):
(1) (A fluoroscopic study; known; as an upper gastrointestinal series)
(2) (caution with non water soluble contrast; is; mandatory as the usage of barium)
(3) (as the usage; of barium can impede; surgical revision and lead)
(4) ( ; to increased; post operative complications)

Supervised-OIE (using discourse-aware TS framework for preprocessing):
(5) #1 0 (A fluoroscopic study; is; typically, the next step in management)
(5a)        L:ELABORATION   #2
(5b)        L:CONTRAST      #3
(6) #2 1 (This; fluoroscopic study is known; as an upper gastrointestinal series)
(7) #3 0 (Caution with non water soluble; is; mandatory)
(7a)        L:CONTRAST      #1
(7b)        L:CONDITION     #7
(7c)        L:BACKGROUND    #4
(7d)        L:BACKGROUND    #5
(7e)        L:BACKGROUND    #6
(8) #4 1 (The usage of barium; can impede; surgical revision)
(8a)        L:LIST          #5
(8b)        L:LIST          #6
(9) #5 1 (The usage of barium; can lead; to increased post operative complications)
(9a)        L:LIST          #4
(9b)        L:LIST          #6
(10) #6 1 (The usage of barium; to increased; post operative complications)
(10a)        L:LIST         #4
(10b)        L:LIST         #5
(11) #7 1 (Volvulus; is suspected; )
\end{BVerbatim}
\end{adjustbox}
\caption{Comparison of the propositions extracted by Supervised-OIE \cite{stanovsky2018supervised} \textit{with} (5-11) and \textit{without} (1-4) using our discourse-aware TS approach as a preprocessing step.}
  \label{fig:ComparativeAnalysisSystems_supervisedOIE}
\end{figure*}


\section{Usage}
\textsc{DisSim} can be either used as a Java API, imported as a Maven dependency, or as a service which we
provide through a command line interface or a REST-like web service that can be deployed via docker.
It takes as input NL text in the form of a single sentence. Alternatively, a file containing a sequence of sentences can be loaded. 
The result of the transformation process is either written to the console or stored in a specified output file in JSON format. 
We also provide a browser-based user interface, where the user can directly type in sentences to be processed (see Figure \ref{screenshot_UI}).
\footnote{A demonstration video is available online: \url{https://streamable.com/08clo}.}

\section{Experiments}
For the English version, we performed both a thorough manual analysis and automatic evaluation across three commonly used TS datasets from two different domains in order to assess the performance of our framework with regard to the sentence splitting subtask. The results show that our proposed sentence splitting approach outperforms the state of the art in structural TS, returning fine-grained simplified sentences that achieve a high level of grammaticality and preserve the meaning of the input. The full evaluation methodology and detailed results are reported in \newcite{niklaus-etal-2019-transforming}.
In addition, a comparative analysis with the annotations contained in the RST Discourse Treebank \cite{carlson2002rst} demonstrates that we are able to capture the contextual hierarchy between the split sentences with a precision of almost 90\% and reach an average precision of approximately 70\% for the classification of the rhetorical relations that hold between them. 
The evaluation of the German version is in progress.

\section{Application in Downstream Tasks}
An extrinsic evaluation was carried out on the task of Open IE \cite{Banko07}. It revealed that when applying \textsc{DisSim} as a preprocessing step, the performance of state-of-the-art Open IE systems can be improved by up to 346\% in precision and 52\% in recall, i.e. leading to a lower information loss and a higher accuracy of the extracted relations. For details, the interested reader may refer to \newcite{niklaus-etal-2019-transforming}.

Moreover, most current Open IE approaches output only a loose arrangement of extracted tuples that are hard to interpret as they ignore the context under which a proposition is complete and correct and thus lack the expressiveness needed for a proper interpretation of complex assertions \cite{niklaus-etal-2018-survey}. As illustrated in Figure \ref{fig:ComparativeAnalysisSystems_supervisedOIE}, with the help of the semantic hierarchy generated by our discourse-aware sentence splitting approach the output of Open IE systems can be easily enriched with contextual information that allows to restore the semantic relationship between a set of propositions and, hence, preserve their interpretability in downstream tasks.

\section{Conclusion}
We developed and implemented a discourse-aware syntactic TS approach that recursively splits and rephrases complex English or German sentences into a semantic hierarchy of simplified sentences.
The resulting lightweight semantic representation can be used to facilitate and improve a variety of AI tasks.

\bibliography{acl2019}
\bibliographystyle{acl_natbib}

\end{document}